\documentclass[conference]{IEEEtran}
\IEEEoverridecommandlockouts

\usepackage{subfigure}
\usepackage[table,xcdraw,dvipsnames]{xcolor}
\usepackage{multirow}
\usepackage{mathtools}
\usepackage{outlines}
\usepackage{times}
\usepackage{multicol}
\usepackage{breqn}
\usepackage[colorlinks=true,linkcolor=black,anchorcolor=black,citecolor=black,filecolor=black,menucolor=black,runcolor=black,urlcolor=black]{hyperref}
\usepackage{amsmath,amsthm,amssymb,accents,mathtools,amsfonts} 
\usepackage{algorithm}
\usepackage{algorithmic}
\usepackage{makecell}
\usepackage[font=footnotesize]{caption}
\usepackage{subcaption}
\usepackage[bb=boondox]{mathalfa}
\usepackage{graphicx}
\usepackage{epsfig}
\usepackage{authblk}
\usepackage{footnote}
\usepackage{booktabs}
\usepackage{enumitem}
\setenumerate[1]{label=\Roman*.}
\setenumerate[2]{label=\Alph*.}
\setenumerate[3]{label=\arabic*.}
\setenumerate[4]{label=\roman*.}

\begin{document}

\newif\ifcommentson
\commentsontrue
\newcounter{BohaoCount}
\addtocounter{BohaoCount}{1}
\newcommand{\bohao}[1]{\textcolor{Green}{\ifcommentson\textbf{(\theBohaoCount) BZ}: \textbf{(#1)}\fi}\addtocounter{BohaoCount}{1}}

\newcounter{DanielCount}
\addtocounter{DanielCount}{1}
\newcommand{\Dan}[1]{\textcolor{Orange}{\ifcommentson\textbf{(\theDanielCount) DH}: \textbf{(#1)}\fi}\addtocounter{DanielCount}{1}}

\newcounter{RamCount}
\addtocounter{RamCount}{1}
\newcommand{\Ram}[1]{\textcolor{Red}{\ifcommentson\textbf{(\theRamCount) RV-M}: \textbf{(#1)}\fi}\addtocounter{RamCount}{1}}

\newcounter{FixCount}
\addtocounter{FixCount}{1}
\newcommand{\fix}[1]{\textcolor{Purple}{\ifcommentson\textbf{(\theFixCount) FIX}: (#1)\fi}\addtocounter{FixCount}{1}}

\newtheorem{defn}{Definition}
\newtheorem{rem}[defn]{Remark}
\newtheorem{lem}[defn]{Lemma}
\newtheorem{prop}[defn]{Proposition}
\newtheorem{assum}[defn]{Assumption}
\newtheorem{ex}[defn]{Example}
\newtheorem{thm}[defn]{Theorem}
\newtheorem{cor}[defn]{Corollary}
\newtheorem{con}[defn]{Conjecture}
\newtheorem{problem}[defn]{Problem}

\providecommand{\R}{\ensuremath \mathbb{R}}
\providecommand{\IR}{\ensuremath \mathbb{IR}}
\providecommand{\N}{\ensuremath \mathbb{N}}
\providecommand{\Q}{\ensuremath \mathcal{Q}}
\providecommand{\A}{\ensuremath \mathcal{A}}
\providecommand{\U}{\ensuremath \mathcal{U}}
\newcommand{\unitcircle}{\mathbb{S}^1}

\newcommand{\ubar}[1]{\underaccent{\bar}{#1}}

\newcommand{\regtext}[1]{\mathrm{\textnormal{#1}}}
\newcommand{\ol}[1]{\overline{#1}}
\newcommand{\ul}[1]{\underline{#1}}
\newcommand{\defemph}[1]{\emph{#1}}
\newcommand{\ts}[1]{\textsuperscript{#1}}

\newcommand{\comp}{^{\regtext{C}}}
\newcommand{\card}[1]{\left\vert#1\right\vert}
\newcommand{\proj}{\regtext{proj}}
\newcommand{\norm}[1]{\left\Vert#1\right\Vert}
\newcommand{\abs}[1]{\left\vert#1\right\vert}
\newcommand{\pow}[1]{\mathcal{P}\!\left(#1\right)}
\newcommand{\diag}[1]{\regtext{diag}\!\left(#1\right)}
\newcommand{\eig}[1]{\regtext{eig}\!\left(#1\right)}
\newcommand{\union}{\bigcup}
\newcommand{\intersection}{\bigcap}
\newcommand{\trans}{^\top}
\newcommand{\inv}{^{-1}}
\newcommand{\pinv}{^{\dagger}}
\newcommand{\sign}{\regtext{sign}}
\newcommand{\expm}{\regtext{exp}}
\newcommand{\logm}{\regtext{log}}
\newcommand{\skw}{_{\times}}
\newcommand{\bigO}{\mathcal{O}}
\newcommand{\bdry}[1]{\regtext{bd}\!\left(#1\right)}
\renewcommand{\ker}[1]{\regtext{ker}\!\left(#1\right)}
\newcommand{\convhull}[1]{\regtext{CH}\!\left(#1\right)}

\newcommand{\lbl}[1]{_{\regtext{#1}}}
\newcommand{\lo}{\lbl{lo}}
\newcommand{\hi}{\lbl{hi}}

\newcommand{\emptyarr}{[\ ]}
\newcommand{\zeros}{\textit{0}}
\newcommand{\ones}{\textit{1}}
\newcommand{\eye}{\regtext{\textit{I}}}

\newcommand{\interval}[1]{[ #1 ]}
\newcommand{\iv}[1]{[ #1 ]}
\newcommand{\nom}[1]{#1}
\newcommand{\setop}[1]{{\mathrm{\texttt{#1}}}}
\newcommand{\lb}[1]{\underline{#1}}
\newcommand{\ub}[1]{\overline{#1}}

\newcommand{\qA}{q_A(t)}
\newcommand{\Wp}{W_{\text{passive}}}

\providecommand{\R}{\ensuremath \mathbb{R}}
\newcommand{\plan}{_p}
\newcommand{\prev}{\lbl{prev}}
\providecommand{\tfin}{t\lbl{f}}

\newcommand{\zi}{z_i}
\newcommand{\zj}{z_j}
\newcommand{\rbf}{\mathbf{r}(t)}

\newcommand{\bH}{H}
\newcommand{\Hq}{H(\q, \theta)}
\newcommand{\Hqdot}{\dot{H}(\q, \theta)}
\newcommand{\Haqt}{H_a(\q, \theta)}

\newcommand{\bC}{C}
\newcommand{\Cq}{C(\q, \qd, \theta)}

\newcommand{\bG}{g}
\newcommand{\Gq}{g(\q, \theta)}

\newcommand{\intparams}{[\theta]}
\newcommand{\nomparams}{\theta_0}
\newcommand{\trueparams}{\theta}

\newcommand{\unom}{u_{\text{nom}}(\qA, \nomparams)}
\newcommand{\urob}{u_{\text{fb}}(\qA)}
\newcommand{\wdist}{w(\qA, \nomparams, \intparams)}

\newcommand{\Gqt}{G(\q)}
\newcommand{\GTqt}{G^T(\q)}

\newcommand{\Kr}{K_r}

\makeatletter
\newcommand{\smalloplus}{\mathbin{\mathpalette\make@small\oplus}}
\newcommand{\smallotimes}{\mathbin{\mathpalette\make@small\otimes}}

\newcommand{\lambdamin}{\lambda_h}
\newcommand{\lambdamax}{\lambda_H}
\newcommand{\sigmin}{\sigma_{h}}
\newcommand{\sigmax}{\sigma_{H}}

\newcommand{\qgoal}{q\lbl{goal}}
\newcommand{\qstart}{q\lbl{start}}

\newcommand{\normrho}{||\rho([\Phi])||}
\newcommand{\normwmax}{||w_M||}
\newcommand{\wmax}{w_M}
\newcommand{\wmaxj}{w_{M, j}}

\newcommand{\J}{J(q(t))}
\newcommand{\JT}{J^T(q(t))}
\newcommand{\Jd}{\dot{J}(q(t))}
\newcommand{\cp}{c(q(t))}
\newcommand{\cd}{\dot{c}(q(t))}
\newcommand{\cdd}{\ddot{c}(q(t))}
\newcommand{\cde}{\J\qd}
\newcommand{\cddeb}{\Jd\qd + \J\qdd}
\newcommand{\q}{q(t)}
\newcommand{\qa}{q_a(t)}
\newcommand{\qu}{q_u(t)}
\newcommand{\qd}{\dot{q}(t)}
\newcommand{\qda}{\dot{q}_a(t)}
\newcommand{\qdu}{\dot{q}_u(t)}
\newcommand{\qdd}{\ddot{q}(t)}
\newcommand{\qdda}{\ddot{q}_a(t)}
\newcommand{\qddu}{\ddot{q}_u(t)}

\newcommand{\dq}{q_d(t)}
\newcommand{\dqa}{q_{d,a}(t)}
\newcommand{\dqu}{q_{d,u}(t)}
\newcommand{\dqd}{\dot{q}_d(t)}
\newcommand{\dqda}{\dot{q}_{d,a}(t)}
\newcommand{\dqdu}{\dot{q}_{d,u}(t)}
\newcommand{\dqdd}{\ddot{q}_d(t)}
\newcommand{\dqdda}{\ddot{q}_{d,a}(t)}
\newcommand{\dqddu}{\ddot{q}_{d,u}(t)}

\newcommand{\dqm}{\dot{q}_{m}(t)}
\newcommand{\ddqm}{\ddot{q}_{m}(t)}
\newcommand{\dqma}{\dot{q}_{m,a}(t)}
\newcommand{\ddqma}{\ddot{q}_{m,a}(t)}

\title{System Identification For Constrained Robots}
\author{Bohao Zhang$^{1,*}$, Daniel Haugk$^{2,*}$,  and Ram Vasudevan$^1$
\thanks{$^{1}$Robotics Institute, University of Michigan, Ann Arbor, MI $\langle$\texttt{jimzhang,ramv} $\rangle$ @umich.edu.}
\thanks{$^{2}$University of Stuttgart \texttt{st161112@stud.uni-stuttgart.de}.}
\thanks{$^{*}$ Authors share equal contribution}
\thanks{This work is supported by the Air Force Office of Scientific Research.}
}
\maketitle
            
\begin{abstract}
    \label{sec:abstract}
 Identifying the parameters of robotic systems, such as motor inertia or joint friction, is critical to satisfactory controller synthesis, model analysis, and observer design. 
Conventional identification techniques are designed primarily for unconstrained systems, such as robotic manipulators. 
In contrast, the growing importance of legged robots that feature closed kinematic chains or other constraints, poses challenges to these traditional methods. 
This paper introduces a system identification approach for constrained systems that relies on iterative least squares to identify motor inertia and joint friction parameters from data.
The proposed approach is validated in simulation and in the real-world on Digit, which is a 20 degree-of-freedom humanoid robot built by Agility Robotics.
In these experiments, the parameters identified by the proposed method enable a model-based controller to achieve better tracking performance than when it uses the default parameters provided by the manufacturer.
The implementation of the approach is available at \href{https://github.com/roahmlab/ConstrainedSysID}{https://github.com/roahmlab/ConstrainedSysID}.
\end{abstract}
\section{Introduction}
\label{sec:intro}

A variety of robotic applications including controller synthesis, model analysis, and observer design, require an accurate dynamics model.
Though manufacturers of robots try to provide such a model, manufacturing or operation variability sometimes mean that these models are inaccurate.
To address this challenge, prior literature has devised system identification methods to compute inertial parameters and friction parameters for actuated joints of a robot from data\cite{Gautier2011, Han2020, Janot2014AGI}.
Though powerful, these traditional system identification methods have focused on unconstrained systems (e.g., robots without closed loop kinematic chains).
This is a notable gap as many modern humanoid robots contain closed kinematic chains like Digit \cite{paper-agility}, Atlas \cite{paper-atlas}, or Optimus \cite{paper-teslabot}. 
In particular, these humanoid robots will be deployed in safety critical situations around humans. 
Traditionally to ensure safe operation, roboticists have relied upon model-based control which requires an accurate dynamical model of the robot.

\begin{figure}[t]
    \centering
    \includegraphics[width=\columnwidth]{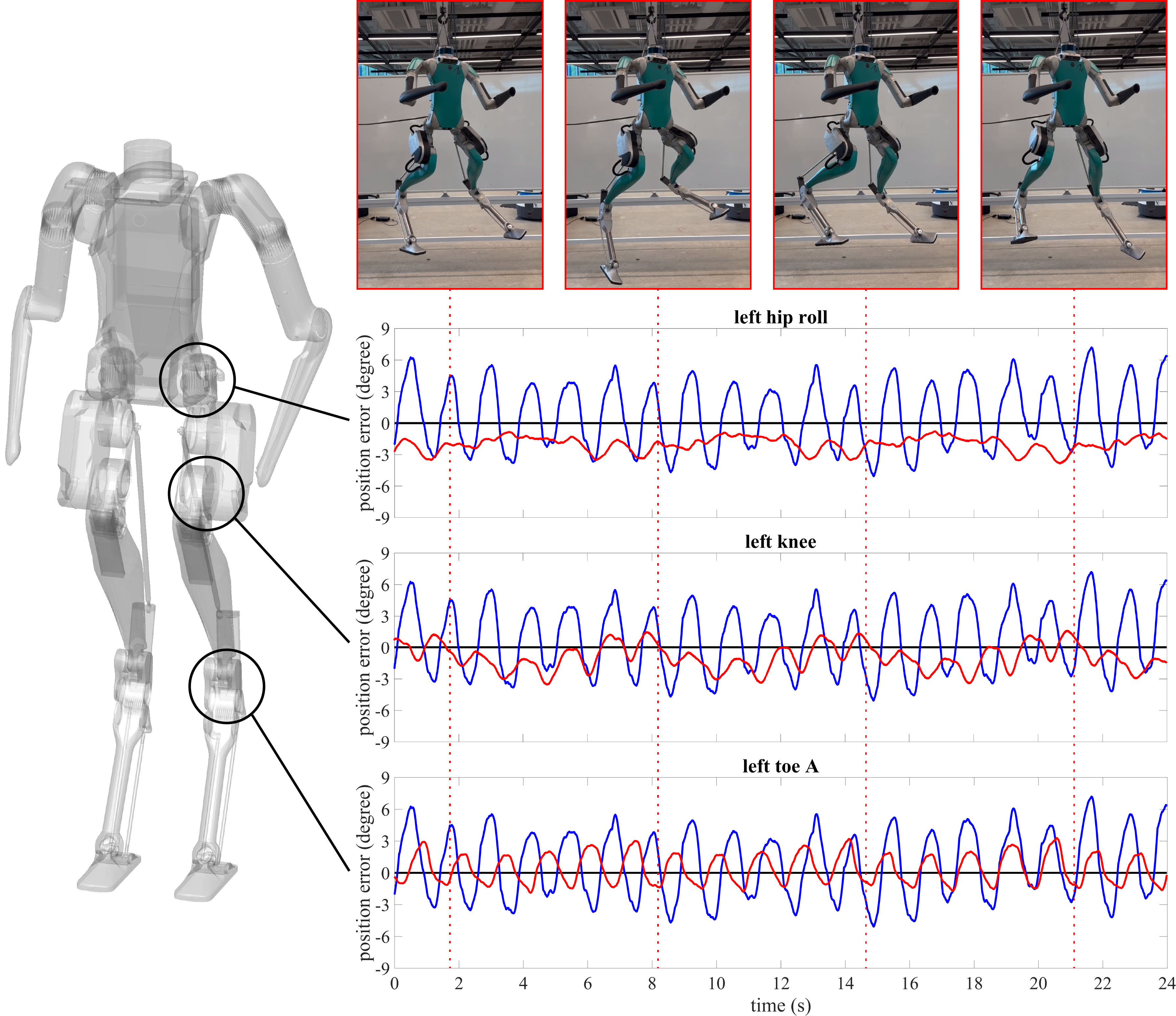}
    \caption{
    This paper proposes a system identification method for systems with constraints such as the humanoid, Digit (image on left).
    Note that Digit has a closed loop kinematic constraint in each of its legs. 
    By using the algorithm developed in this paper, one can identify the inertial and friction parameters of the system from data. 
    To illustrate the utility of this method, this paper compared the performance of a model based tracking controller to track a user-specified trajectory. 
    The tracking performance of the controller was significantly less when using the parameters identified by the algorithm developed in this paper (drawn in red on the right) when compared to using the parameters specified by the manufacturer (drawn in blue on the right).
    The actual behavior of the robot while following the user-specified trajectory using the parameters identified by the algorithm developed in this paper can be seen on the top right of the image at 4 time instances. 
    }
    \label{fig:summary_figure}
\end{figure}

\subsection{Related Work}

Before describing the contributions of this paper, we briefly summarize related work.
One of the first methods for estimating the inertial parameters of a robotic system relied on least square \cite{An1985}. 
Though computationally straightforward, this method faced issues due to the rank deficiency of the regressor matrix from linearly dependent and unidentifiable inertial parameters. 
Later, more advanced methods \cite{Gautier2011, Han2020, Janot2014AGI} were developed to deal with these issues.
The rank deficiency problem was addressed by applying numerical \cite{Gautiera} or symbolic \cite{Gautier1990, Khalil1994} methods to calculate a minimal set of inertial parameters.
This process, known as regrouping, resulted in a sub-regressor matrix that allowed for a more reliable parameter estimation \cite{Han2020, Venture2009}.
Despite these advancements, least squares-based methods have further limitations, including sensitivity to noise and the potential to generate physically inconsistent parameters (e.g., negative masses or friction coefficients and negative-definite inertia tensors).
Subsequently a Linear Matrix Inequality (LMI) constraint was introduced on the inertial parameters to ensure that they were physically realizable  \cite{Sousa2014, Sousa2019,Wensing2018}.

Traditional system identification techniques, although effective for systems with open kinematic trees \cite{paper-tree-structure}, struggle with closed kinematic chains due to constraints on their equations of motion.
To begin addressing this problem, a method has been developed to identify a minimal representation of the dynamics for systems with closed-loop kinematic chain constraints \cite{Bennis1992CalculationOT}. 
This technique has been extended to perform system identification by eliminating some of the inertial parameters by utilizing the representation of the constraint that describes the closed-loop kinematic constraint\cite{Tan2022}.
Unfortunately this method is only applicable for systems with constraint representations that are linear, which renders it inapplicable for performing system identification for modern humanoids. 
Other approaches have tried to perform system identification for nonlinear constraint representations \cite{Shome_1998}; however, these methods are unable to enforce constraints on the non-negativity of inertial parameters, which can result in non-physically realizable results.

\subsection{Contributions and Organization}

The contributions of this paper are two-fold.
First, we propose an optimization-based technique for system identification for constrained robotic systems from data that is able to guarantee the physical realizability of any estimate parameters. 
Second, as depicted in Figure \ref{fig:summary_figure}, we illustrate that our proposed method can be applied to identify model parameters for real-world systems that can lead to superior performance when compared to using manufacturer specified parameters. 
In particular, we illustrate that when we apply inverse dynamics control to track trajectories on the $20$ degree-of-freedom humanoid robot, Digit, the parameters identified by our algorithm lead to superior performance when compared to the manufacturer specified parameters.

The remainder of this paper is organized as follows. 
Section \ref{sec:preliminaries} introduces  mathematical concepts used throughout the paper. 
Section \ref{sec:sysid} describes the proposed method for system identification of fully actuated constrained systems. 
Section \ref{sec:experimental} describes multiple validation experiments and demonstrates the performance the proposed algorithm in the real-world on Digit.

\section{Preliminaries}
\label{sec:preliminaries}

This section formally describes the equations of motion of constrained systems, formulates these equations using a regressor matrix, and concludes by making an observation related to fully actuated systems.

\subsection{Equations of Motion of Constrained Systems}
\label{ssec:eom}

Consider a robotic system with $n$ joints.
The generalized trajectory is denoted as $q:[0,\infty)\in\Q$, where $\Q \subset \R^{n}$ is the robot configuration space.
The number of actuated joints (motors) and unactuated joints is given by $n_a$ and $n_u$, respectively, with $n_a + n_u = n$. 
The trajectory of actuated and unactuated joints are denoted as $\qa\in\R^{n_a}$ and $\qu\in\R^{n_u}$, respectively, which are entries from $\q$.
Denote the set of indices for all actuated joints and unactuated joints as $\A\subset\N$ and $\U\subset\N$, respectively.
Thus, $\A \cap \U = \emptyset$ and $\A \cup \U = \{1,\ldots,n\}$.

The \defemph{standard dynamic parameters} of the robot are given by a vector $\theta \in \R^{14n}$ that is composed of two parts.
The first part of $\theta$ is denoted by $\theta_{\text{ip}} \in \R^{10n}$ and is made up of
        \begin{multline}
            \theta_{\text{ip},j}=[XX_j, XY_j, XZ_j, YY_j, YZ_j, ZZ_j, \\
            p_{x,j}m_j, p_{y,j}m_j,p_{z,j}m_j, m_j]^T,
        \end{multline}
which are the inertial parameters of link $j$.
The first six components of $\theta_{\text{ip},j}$ are the inertia tensor, the next three components of $\theta_{\text{ip},j}$ are the first moments of area, and the last component of $\theta_{\text{ip},j}$ is the mass.
Note, $p_j = [p_{x,j}, p_{y,j}, p_{z,j}]^T$ is the center of mass, for each link $j \in \{1,\ldots,n\}$.
The second part of $\theta$ is denoted by $\theta_{\text{f}} \in \R^{4n}$ is a vector of parameters related to joint friction and is made up of
    \begin{equation}
        \theta_{\text{f},j} =[F_{c_{j}}, F_{v_j}, I_{a_j}, \beta_j]^T
    \end{equation}
as the friction parameters for each joint $j$. 
We describe how these parameters are used in \eqref{eq:friction}.

Because the focus of this paper is on robots with constraints, we assume that there exists a constraint function that is denoted by $c:\Q\rightarrow\R^{n_c}$, where $n_c\geq0$ is the number of constraints.
The constraints are satisfied when $c(\q) = \mathbb{0}_{n_c}$.
The Jacobian of $c$ at $\q$ is denoted by $J(\q)\in\R^{n_c\times n}$.

The equation of motion for the system is given by:
\begin{multline} \label{eq:CEoM}
    H(\q, \theta) \qdd + C(\q, \qd, \theta) \qd + g(q, \theta) + \\
    + F(\qd, \qdd, \theta) = \tau(t) + \JT \lambda(t),
\end{multline}
while satisfying the following constraints for all time $t \geq 0$:
\begin{align}
    &c(q(t)) = \mathbb{0}_{n_c} \label{eq:cons-pos} \\
    &\dot{c}(q(t)) = J(q(t))\qd = \mathbb{0}_{n_c} \label{eq:cons-vel} \\
    &\ddot{c}(q(t)) = J(q(t))\qdd + \dot{J}(q(t))\qd = \mathbb{0}_{n_c}, \label{eq:cons-acc}
\end{align}
where $H(\q, \theta) \in \R^{n\times n}$ is the positive definite \defemph{inertia mass matrix},
$C(\q, \qd, \theta) \in \R^{n\times n}$ is the \defemph{Coriolis matrix}, 
$g(q, \theta) \in \R^{n}$ is the \defemph{gravitational force vector}, 
and $\lambda(t)\in\R^{n_c}$ is the reaction forces/wrenches to maintain the constraints at time $t$ \cite[Section 8.1]{featherstone2014rigid}.
The joint friction forces are  $F(\qd, \qdd, \theta) \in \R^{n}$, which are modeled by generalizing \cite[(2)]{Han2020}:
\begin{multline} \label{eq:friction}
    F_j(\Dot{q}_j(t), \ddot{q}_j(t), \theta_{\text{f},j}) = F_{c_j} \, \text{sign}(\Dot{q}_j(t)) + \\
    + F_{v_j} \, \Dot{q}_j(t) + I_{a_j} \, \ddot{q}_j(t) + \beta_j,
\end{multline}
where $F_{c_j}$ is the \defemph{static friction coefficient}, $F_{v_j}$ is the \defemph{viscous friction coefficient}, $I_{a_j}$ is the \defemph{transmission inertia}, and $\beta_j$ is the \defemph{offset/bias term} for each joint $j$.
The torque applied on each joint at time $t$ denoted by $\tau(t)\in\R^{n}$ is given by
\begin{equation}
    \tau(t) = Bu(t),
\end{equation}
where $u(t)\in\R^{n_a}$ is the control input at time $t$.
The \defemph{transmission matrix}, $B\in\R^{n\times n_a}$,  which is defined as 
\begin{equation}
    B_u = \mathbb{0}_{n_u\times n_a},\quad 
    B_a = \mathbb{1}_{n_a\times n_a},
\end{equation}
where $B_a$ and $B_u$ are the collection of all actuated rows and all unactuated rows of $B$, respectively.

\subsection{Regressor Matrix}
\label{ssec:roip}

Our objective is to identify the standard dynamics parameters, $\theta$, from data. 
To do this, first note that the left hand side of \eqref{eq:CEoM} can be rewritten as a linear function of the standard dynamics parameters: 
\begin{equation} \label{eq:W}
    W(\q, \qd, \qdd)\theta = \tau(t) +  \JT \lambda(t),
\end{equation}
with $W(\q, \qd, \qdd)\in\R^{n\times 14n}$ \cite{paper-regressor}.
Second, because some parameters are linearly dependent or unidentifiable \cite{Gautier1990, Khalil1994}, the left side can be further simplified to
\begin{equation} \label{eq:Y}
    Y(\q, \qd, \qdd)\pi = \tau(t) +  \JT \lambda(t),
\end{equation}
where we refer to $Y(\q, \qd, \qdd) \in \R^{n\times n_{id}}$ as the \defemph{regressor matrix}, which is composed of the maximum number of linear independent columns of $W(\q, \qd, \qdd)$, and $\pi \in \R^{n_{id}}$ is defined as the \defemph{base parameters}, which is the vector of the collection of corresponding entries in $\theta$.
Denote the linearly dependent columns of $W(\q, \qd, \qdd) \in \R^{n \times n_d}$ as $W_d(\q, \qd, \qdd)$ and the corresponding entries of $\theta$ as $\theta_d \in \R^{n_d}$, where $n_d$ is the number of linear dependent columns and $n_{id} + n_d = n$.

There exists the following linear relationships \cite[(32),(33)]{Sousa2014}:
\begin{align}
    Y(\q, \qd, \qdd) &= W(\q, \qd, \qdd)P_b \\
    W_d(\q, \qd, \qdd) &= W(\q, \qd, \qdd)P_d \\
    W_d(\q, \qd, \qdd) &= Y(\q, \qd, \qdd)K_d,
\end{align}
where $P_b$ and $P_d$ are two mutually exclusive parts of a permutation matrix.
If one is given access to noise-free data, then one can compute $P_b$, $P_d$ and $K_d$ using a rule-based technique \cite{Gautier1990, paper-regressor}. 
In the presence of noise, one can compute $P_b$, $P_d$ and $K_d$ using a numerical technique such as QR decomposition or singular value decomposition \cite{Gautiera}.
As a result, the relationship between the base parameters, $\pi$, the dependent parameters, $\theta_d$. and the standard dynamics parameters,  $\theta$, can be summarized as follows \cite[(33), (38), (49)]{Sousa2014}:
\begin{align} 
    \theta_d &= P_d^T\theta \\
    \pi &= (P_b^T + K_dP_d^T)\theta \\
    \theta &= P_b(\pi - K_d\theta_d) + P_d\theta_d. \label{eq:pi-to-theta}
\end{align}

\subsection{Fully-Actuated Representation}
\label{ssec:full-actuated}

When the unactuated dimension, $n_u$, is equal to the number of constraint,s $n_c$, the system is \defemph{fully-actuated} \cite[Section 8.12]{featherstone2014rigid}.
The focus of this paper is to perform system identification for constrained, fully actuated systems. 
To accomplish this objective, we describe how to rewrite the system dynamics in an equivalent fashion, which we refer to as the \defemph{fully-actuated representation}, that can facilitate system identification and control of robots with constraints. 
This is because this representation enables us to describe the unactuated joints trajectory as a function of the actuated joints trajectory.

Before formally describing how to construct this fully-actuated representation, we make the following assumption:
\begin{assum} \label{assum-uniqueIK}
    For any actuated joint positions $\forall\qa\in\Q_a$, there exists one and only one set of unactuated joint positions $\qu\in\Q_u$ such that the constraints are satisfied, i.e., $c(\q) = \mathbb{0}_{n_c}$.
\end{assum}
\noindent This assumption ensures that the constraint Jacobian $J_u(\q) \in \R^{n_c\times n_u}$, 
is always invertible according to the inverse function theorem \cite[Theorem 5.2.1]{JerryShurman}.
Note in the experimental section, we numerically evaluate whether this assumption is satisfied.
We denote the $\qu$ that satisfies the constraints as a function of $\qa$:
\begin{equation} \label{eq-IK}
    \qu = \Gamma(\qa).
\end{equation}

Using Assumption \ref{assum-uniqueIK}, one can prove the following theorem whose proof can be found in Appendix \ref{app-thm-fully-actuated}: 
\begin{thm} \label{thm-fully-actuated}
    Suppose at time $t$, the actuated joint position, velocity, and acceleration are $\qa$, $\qda$, and $\qdda$, respectively. 
    Let $G(\q)\in\R^{n\times n_a}$ be a matrix whose unactuated rows are defined as
    \begin{equation}
        G_u(\q) = -J_u^{-1}(\q) J_a(\q),
    \end{equation}
    and whose actuated rows are defined as
    \begin{equation}
        G_a(\q) = \mathbb{1}_{n_a \times n_a},
    \end{equation}
    where $J_u(\q) \in \R^{n_c\times n_u}$ and $J_a(\q) \in \R^{n_c\times n_a}$ are the collection of unactuated and actuated columns of the constraint Jacobian $J(\q)$, respectively.
    Then the velocity and acceleration of all joints at time $t$ are
    \begin{align} 
        &\qd = G(\q)\qda \label{eq-fill-full-joints1} \\
        &\qdd = G(\q)\qdda + \dot{G}(\q)\qda, \label{eq-fill-full-joints2}
    \end{align}
    and the control input at time $t$ is
    \begin{equation} \label{eq-constrainedID-torque}
        u(t) = G^T(\q)Y(\q, \qd, \qdd)\pi. 
    \end{equation}
\end{thm}
\section{System Identification For Fully-Actuated Constrained Systems}
\label{sec:sysid}

For system identification, assume that one has access to $N > 0$ measurements of the system state.
Suppose the sequences of position, velocity, and acceleration measurements are denoted as:
\begin{align}
    \mathbf{q}_{1:N} &= \{q(t_i)\}_{i=1}^N \\
    \dot{\mathbf{q}}_{1:N} &= \{\dot{q}(t_i)\}_{i=1}^N \\
    \ddot{\mathbf{q}}_{1:N} &= \{\ddot{q}(t_i)\}_{i=1}^N,
\end{align}
where $t_i \geq 0$ for $i\in\{1,\ldots,N\}$ are the time instances of each measurement.

Let the \defemph{constrained regressor matrix} at each time instance be defined as
\begin{equation}
    GY_i = G^T(q(t_i))Y(q(t_i), \dot{q}(t_i), \ddot{q}(t_i))
\end{equation}
for $i\in\{1,\ldots,N\}$.
We then can build the \defemph{constrained observation matrix} $\mathbf{GY}$ as follows:
\begin{equation} \label{eq:observation-regressor}
    \mathbf{GY}(\mathbf{q}_{1:N}, \dot{\mathbf{q}}_{1:N}, \ddot{\mathbf{q}}_{1:N}) \vcentcolon= \begin{bmatrix}
        GY_1 \\ 
        GY_2 \\ 
        \vdots \\ 
        GY_m
    \end{bmatrix} \in \R^{n_a N\times n_{id}} .
\end{equation}
The \defemph{observation response} is the collection of all control inputs:
\begin{equation} \label{eq:observation-response}
    \mathbf{U} \vcentcolon= \begin{bmatrix}
        u(t_1) \\ 
        u(t_2) \\ 
        \vdots \\ 
        u(t_m)
    \end{bmatrix} \in \mathbb{R}^{n_a N }.
\end{equation}
To ensure physical consistency of the inertial parameters, the following LMI constraints \cite{Sousa2019} should be satisfied:
\begin{equation} \label{lmi}
    \text{LMI}_j(\theta) \vcentcolon= 
    \begin{bmatrix}
    \left(\frac{\text{tr}(L_j)}{2}\mathbb{1}_{3 \times 3} - L_j\right) \ & p_jm_j \\ p_j^Tm_j & m_j
    \end{bmatrix} \succ \mathbb{0}_{4\times 4},
\end{equation}
where $\text{tr}(\cdot)$ is the trace operator and $L_j$ is the \defemph{shifted inertia tensor} defined as 
\begin{equation}
    L_j \vcentcolon= \begin{bmatrix}
        XX_j & XY_j & XZ_j \\
        XY_j & YY_j & YZ_j \\
        XZ_j & YZ_j & ZZ_j
    \end{bmatrix}
\end{equation}
for each link $j \in \{1,\ldots,n\}$.

To identify the base parameters, $\boldsymbol{\pi}$, from data one can solve the following optimization problem:
\begin{align} 
        &\min_{\pi, \theta_d} && \lVert \mathbf{GY}(\mathbf{q}_{1:N},\dot{\mathbf{q}}_{1:N},\ddot{\mathbf{q}}_{1:N})\pi - \mathbf{U} \rVert_2^2 \label{eq:opt1} \\
        &\text{s.t. } && F_{c_j} \geq 0, \ &\forall j \in \{1,\ldots,n\} \nonumber \\
        & &&F_{v_j} \geq 0, \ &\forall j \in \{1,\ldots,n\} \nonumber  \\
        & &&I_{a_j} \geq 0, \ &\forall j \in \{1,\ldots,n\} \nonumber  \\
        & &&\text{LMI}_j(\theta) \succ \mathbb{0}_{4\times 4}, \ &\forall j \in \{1,\ldots,n\},\nonumber  \\
        & &&\theta = P_b(\pi - K_d\theta_d) + P_d\theta_d & \nonumber 
\end{align}
\noindent To robustly deal with noise and outliers, one can apply normalization and weighting of the observation matrix \eqref{eq:observation-regressor} and the observation response \eqref{eq:observation-response}. 
We do this by using the iterative weighted least squares approach proposed in \cite[Figure 5]{Han2020} to refine our identification process and improve accuracy.
The solution to \eqref{eq:opt1} is denoted as $\pi_0$ and $\theta_{d,0}$.
One can use this solution to compute $\nomparams$ by using \eqref{eq:pi-to-theta}.

\section{Experiments}
\label{sec:experimental}

We tested our method on Digit, which was mounted using a rope with its torso fixed in the air.
We analyzed the performance of the system identification process by cross-validating on a variety of ground-truth datasets.
We also used the identified parameters to perform tracking control of a family of desired trajectories by using an inverse dynamics controller. 
The complete results can be found in our online supplementary material \cite{supp}.

\subsection{Digit Overview}
\label{ssec:digitmodel}

Digit is a humanoid robot with 42 degrees of freedom, developed by Agility Robotics.
This work assumes that the base of Digit is fixed in the air, with no contact between its feet and the ground.
Due to this assumption, the dynamics of each limb do not affect one another.
As a result, we perform the experiments on each limb separately.

\begin{figure}[t]
    \centering
    \includegraphics[width=0.9\columnwidth]{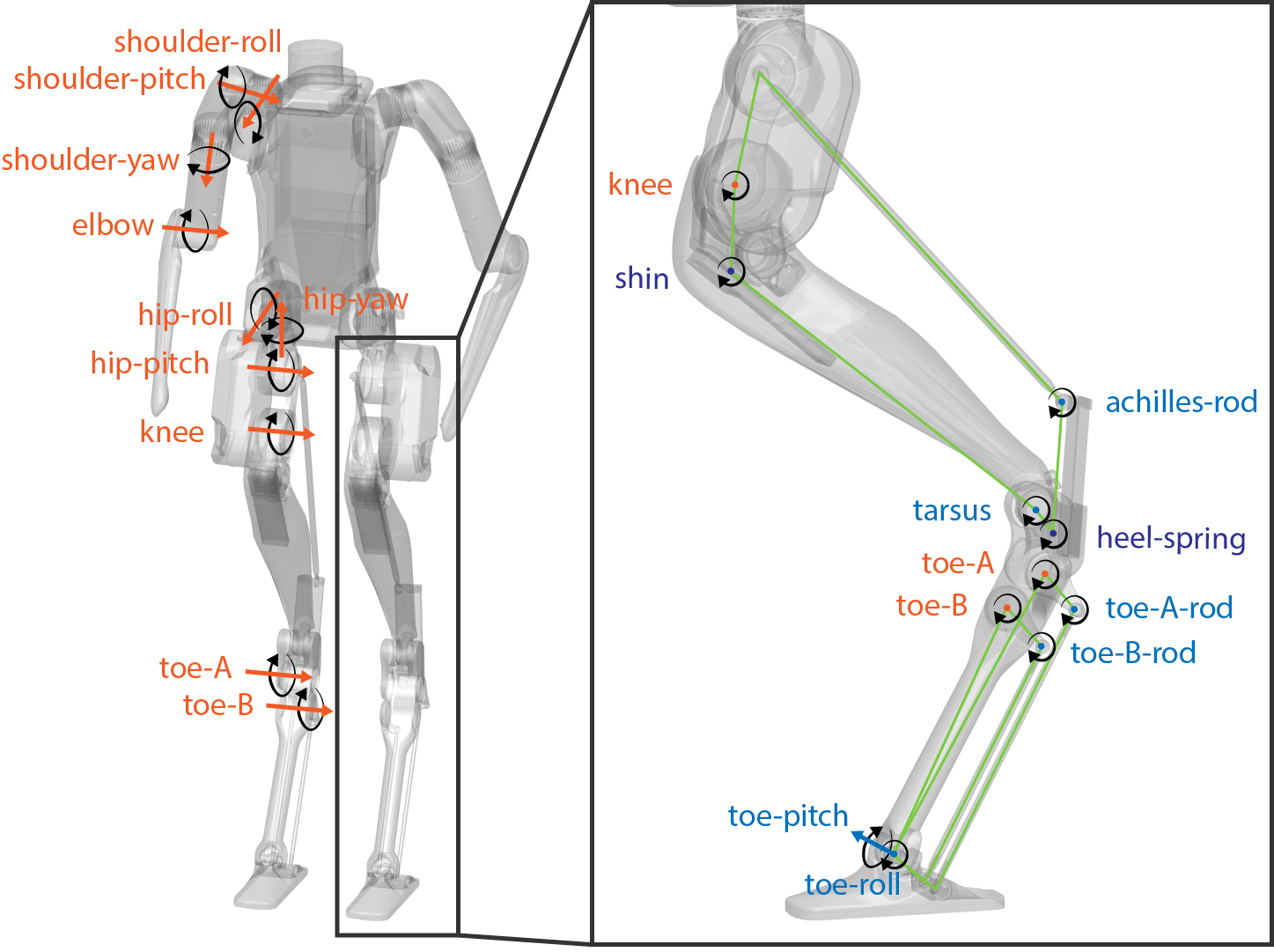}
    \caption{This illustrates the closed-loop structure on the legs of Digit. 
    The orange arrows show the rotation axes of all actuated joints (motors). 
    The blue arrows show the rotation axes of all unactuated joints. 
    The purple arrows show the rotation axes of all springs (shin and heel-spring), 
    which are assumed to be fixed in this paper (see Assumption \ref{ass: sp}). 
    The green lines show how these joints are connected to be closed-loops.}
    \label{fig:digit}
\end{figure}

Digit includes 4 actuated revolute joints per arm and 14 revolute joints per leg, of which 6 are actuated and 8 are passive.
Each leg has 3 closed kinematic chains. The upper leg chain consists of 1 actuated joint (knee) and 4 passive joints (shin, heel-spring, achilles-rod, tarsus). 
The lower leg chains include 2 actuated joints (toe-A, toe-B) and 4 passive joints (toe-pitch, toe-roll, toe-B-rod, toe-B-rod).
The kinematic structure is summarized in Figure \ref{fig:digit}.
All actuated joints have encoders to measure joint position, joint velocity, and motor torque. 
There are also joint encoders on the tarsus.
The low-level control API from Agility Robotics provides access to joint position and velocity measurements on toe-pitch and toe-roll through its internal inverse kinematics algorithm.

We make the following simplifications for Digit.
The springs on Digit's legs exhibit very little movement during actual hardware experiments.
In fact the most recent bipedal robot models (\cite{paper-atlas, paper-teslabot, paper-unitree, paper-fourier-intelligence}), including the latest version of Digit \cite{paper-agility-v4}, do not incorporate such springs.
As a result, we make the following assumption:
\begin{assum} \label{ass: sp}
    All four springs (shin and heel-spring on both legs) on Digit are fixed.
\end{assum}

\noindent By fixing all the springs, each chain in the legs becomes a four-bar linkage mechanism with 3 kinematics constraints.
This results in a total of 15 joints per leg with 9 passive joints and 9 constraints.

We numerically verified that Assumption \ref{assum-uniqueIK} was satisfied by sampling $100$ different actuated joint positions within the corresponding joint limits and solving inverse kinematics for unactuated joints.
In particular, the solutions were always unique.
This finally leads to the following conclusion:
\begin{rem}
    Under Assumptions \ref{ass: sp}, Digit is fully actuated because the number of passive joints $n_{\text{u}}$ is equal to the number of constraints $n_{\text{c}}$.
\end{rem}

\subsection{Implementation Details}

Our system identification method was implemented in MATLAB while our inverse dynamics controller was implemented in C++.
We ran the system identification process and other experiments on a laptop with an Intel-i7 CPU and 32GB RAM.
For legs, the algorithm focused on identifying only motor and friction parameters, as we assumed that the inertial parameters provided by the robot manufacturer were accurate. 
For arms, we identified all the dynamics parameters.
The implementation of the approach is available at \href{https://github.com/roahmlab/ConstrainedSysID}{https://github.com/roahmlab/ConstrainedSysID}.

The system identification process began with collecting data. 
Multiple optimizations were performed offline to generate a set of trajectories for generalization, minimizing the condition number of the observation matrix, as described in \cite{Han2020, Venture2009}. 
These trajectories were tracked on Digit using a PD controller to take measurements of the joint positions, velocities, and motor torques of the actuated joints.
The data processing procedure is as follows:
\begin{enumerate}
    \item[1.] Estimate the acceleration $\qdd$ by applying a $4^{\text{th}}$ order central difference filter to velocity data.
    \item[2.] For the legs, calculate the positions, velocities, and accelerations of the unactuated joints using \eqref{eq-IK}, \eqref{eq-fill-full-joints1}, and \eqref{eq-fill-full-joints2}.
    \item[3.] Apply a fourth-order Butterworth filter with zero delay to the velocity, acceleration, and torque data.
    \item[4.] Downsample the data by a factor of 10 to reduce the number of data points used to perform system idenitification. This reduces the size of the optimization problem which allows for more rapid computation. 
    \item[5.] Compute the regressor matrix at each time instance using a regrouping methods based on QR decomposition \cite[(13)]{Gautiera}.
    \item[6.] Compute $\Gqt$ and then the constrained regressor matrix at each time instance.
    \item[7.] Collect the constrained regressor matrices and the control inputs at all time instances to get the constrained observation matrix and the observation response.
\end{enumerate}
The constrained observation matrix and the observation response were then used to solve \eqref{eq:opt1}.
We used Matlab's ``MultiStart'' function together with ``fmincon.''
This randomly samples 100 initial guesses within the bounds of the decision variables, solves the optimization in parallel and chooses the best solution.

\subsection{Parameters Verification}

\subsubsection{Inverse Dynamics Validation}
\begin{figure*}[ht]
    \centering
    \includegraphics[width=1.9\columnwidth]{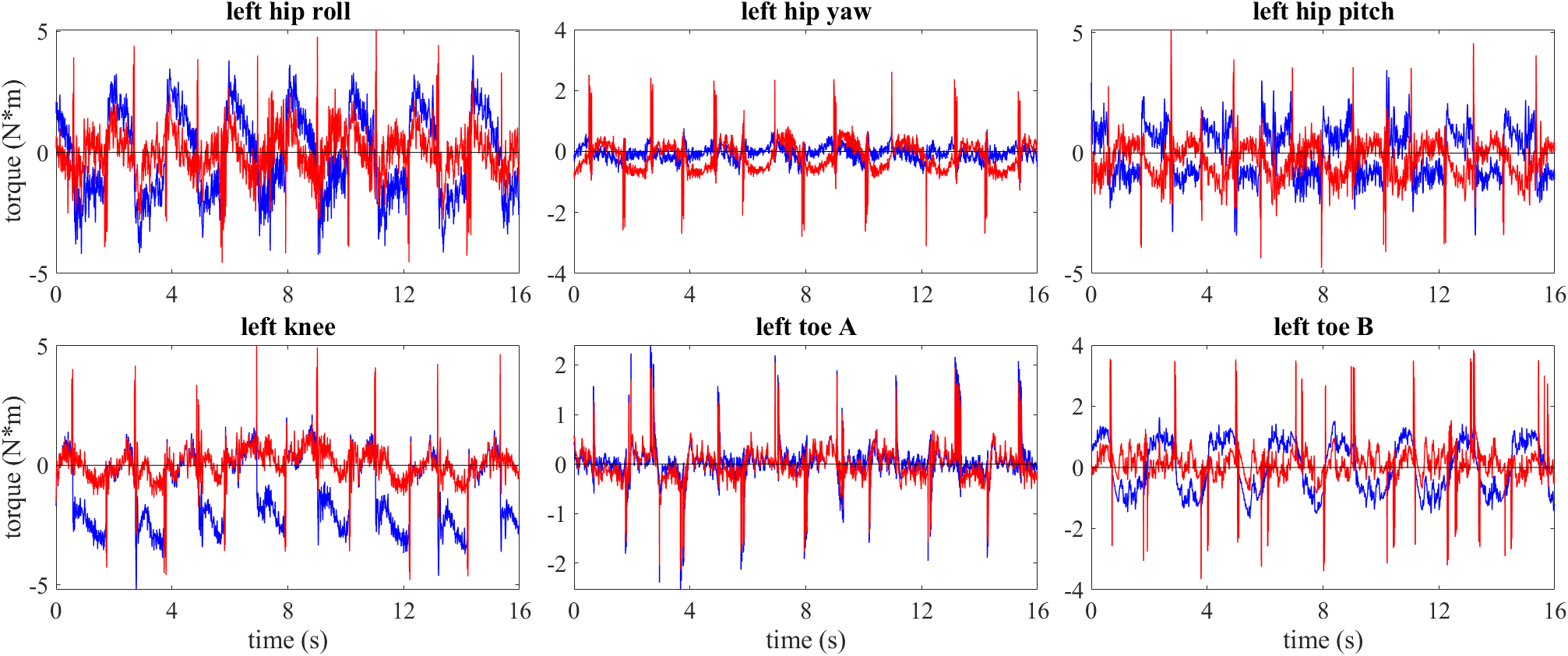}
    \caption{Validation experiment by comparing measured torque and estimated torque on the left leg.
    The \textcolor{red}{\textbf{torque residual based on identified parameters}} is plotted in \textcolor{red}{\textbf{red}}.
    The \textcolor{blue}{\textbf{torque residual based on manufacturer provided parameters}} is plotted in \textcolor{blue}{\textbf{blue}}.
    The data shown is distinct from that used in identification.
    }
    \label{fig:ID_results}
\end{figure*}

To validate the identified parameters, we compared the estimated torque and the measured torque.
A different dataset from the one used for parameter identification was used for this comparison. 
We compute the RMS error for all actuated joints and report the results in Table \ref{tab:ID_results}.
The validation results of the residual between the measured torque and the estimated torque for the left leg are depicted in Figure \ref{fig:ID_results}.
The plots for the right leg and arms can be found in our online supplementary material \cite{supp}.

\begin{table}[t]
    \centering
    \begin{tabular}{c|c|c|c|c}
    \hline
    \multirow{2}{*}{Actuated Joints} & \multicolumn{2}{c|}{Left} & \multicolumn{2}{c}{Right} \\
    \cline{2-5}
     & Ours & Manufacturer & Ours & Manufacturer \\
    \hline
    hip-roll  & \textbf{0.9982} & 1.8116 & \textbf{1.5160} & 1.9214 \\
    hip-yaw   & 0.5642 & \textbf{0.3830} & \textbf{0.5650} & 0.6003 \\
    hip-pitch & \textbf{0.8954} & 1.0983 & 1.2395 & \textbf{1.1012} \\
    knee      & \textbf{0.7384} & 1.8219 & 2.3843 & \textbf{2.0263} \\
    toe-A     & \textbf{0.3675} & 0.3907 & \textbf{0.4581} & 0.4676 \\
    toe-B     & \textbf{0.6688} & 0.9422 & 1.0609 & \textbf{0.4889} \\
    \hline
    shoulder-roll  & \textbf{0.9556} & 2.1060 & \textbf{1.0699} & 1.4821 \\
    shoulder-pitch & \textbf{1.0628} & 1.2693 & \textbf{0.8826} & 1.2412 \\
    shoulder-yaw   & \textbf{0.5121} & 0.5537 & 0.8742 & \textbf{0.4949} \\
    elbow          & \textbf{1.2189} & 1.4197 & 1.3873 & \textbf{1.3634} \\
    \hline
    \end{tabular}
    \caption{The RMS error in N*m between measured torque and estimated torque based on identified parameters and manufacturer provided parameters.}
    \label{tab:ID_results}
\end{table}

\subsubsection{Forward Integration Prediction Validation}

\begin{figure*}[t]
    \centering
    \includegraphics[width=1.9\columnwidth]{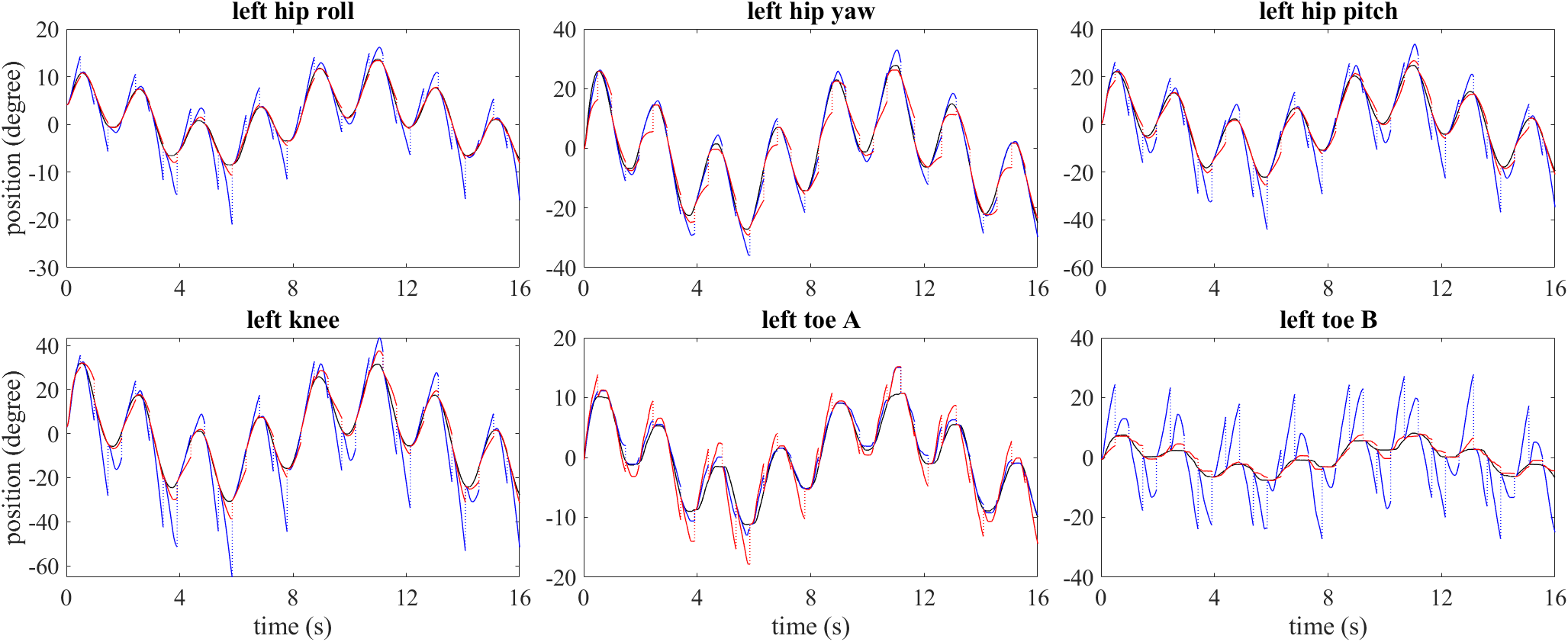}
    \caption{Validation experiment by comparing the estimated trajectory from ode15s forward integration and the ground truth data set on the left leg.
    The \textbf{measured trajectory} is plotted in \textbf{black}.
    The \textcolor{red}{\textbf{estimated trajectory using identified parameters}} is plotted in \textcolor{red}{\textbf{red}}.
    The \textcolor{blue}{\textbf{estimated trajectory using manufacturer provided parameters}} is plotted in \textcolor{blue}{\textbf{blue}}.
    Note that we performed the validation by separating the data set into 31 segments, where the initial condition of the ode15s simulation starts from the beginning of each segment.
    The estimated trajectory between two adjacent segments is connected with vertical dotted lines.
    }
    \label{fig:forward_simulation_results}
\end{figure*}

We also assessed the simulation quality by measuring the discrepancy between the simulation and ground truth data, where the identified parameters are plugged into the system dynamics.
We performed simulations using MATLAB ode15s \cite{paper-ode15s} with the identified parameters plugged into the system dynamics.
We provided the control inputs by applying zero-order hold (ZOH) to the torques recorded from the same validation experiment and compared the difference between the simulated trajectories and the recorded trajectories.
Because we only apply discrete open-loop control, it is inevitable that the simulation eventually diverges from the ground truth.
As a result, we only performed comparison on relatively short horizons.
To be more specific, we separated the data set into 31 segments, each of $0.5$ second length. 
We treated the beginning of each segment of the dataset as the initial condition and simulated forward in ode15s until the end of the segment.
To evaluate the difference between the simulation and the ground truth, we use the $L_2$-norm, which is defined as
\begin{equation}
    L_2\text{-norm} = \sqrt{\sum_{i=1}^{N}|q_{j,\text{simulation}}(t_i)-q_{j,\text{measurement}}(t_i)|^2},
\end{equation}
for each joint $j \in \{1,\ldots,n_q\}$.
The $L_2$ error for all joints over all segments is reported in Table \ref{tab:error_comparison}. 
The plot of trajectories for the left leg is shown in Figure \ref{fig:forward_simulation_results}.
The plots of trajectories for the right leg and arms can be found in our online supplementary \cite{supp}.

\begin{table}[t]
    \centering
    \begin{tabular}{c|c|c|c|c}
    \hline
    \multirow{2}{*}{Actuated Joints} & \multicolumn{2}{c|}{Left} & \multicolumn{2}{c}{Right} \\
    \cline{2-5}
     & Ours & Manufacturer & Ours & Manufacturer \\
    \hline
    hip-roll  & \textbf{0.5497} & 3.1946  & \textbf{1.3118} & 3.4000 \\
    hip-yaw   & 4.1865 & \textbf{2.8339}  & \textbf{3.3326} & 5.0007 \\
    hip-pitch & \textbf{2.4127} & 7.4932  & \textbf{5.9592} & 7.8870 \\
    knee      & \textbf{3.5314} & 11.2491 & 11.8719 & \textbf{11.0496} \\
    toe-A     & 3.0140 & \textbf{1.4248}  & 3.7429 & \textbf{2.7318} \\
    toe-B     & \textbf{1.5589} & 13.9524 & 3.5540 & \textbf{3.0070} \\
    \hline
    shoulder-roll  & \textbf{1.9664} & 9.3720 & \textbf{4.5500} & 7.9586 \\
    shoulder-pitch & \textbf{2.6041} & 4.3640 & \textbf{3.3619} & 5.5375 \\
    shoulder-yaw   & \textbf{2.5248} & 3.5880 & 8.8738 & \textbf{4.1524} \\
    elbow          & \textbf{4.8756} & 7.9837 & \textbf{5.2443} & 7.5329 \\
    \hline
    \end{tabular}
    \caption{$L_2$ error between estimated trajectory from ode15s forward integration and the ground truth data set.}
    \label{tab:error_comparison}
\end{table}

\subsection{Inverse Dynamics Controller Experiment}

To further evaluate the utility of the identified parameters, we perform a trajectory tracking task in the real-world.
To perform trajectory tracking, we considered the following inverse dynamics controller:
\begin{multline}
    u(t) = G^T(\dq)W(\dq, \dqd, \dqdd)\theta_0 + \\
    + K_p e_a(t) + K_d \dot{e}_a(t),
\end{multline}
where $q_d: [0, \infty) \rightarrow \Q$ are the desired trajectories for all joints.
$K_p\in\R^{n_a\times n_a}$ and $K_d\in\R^{n_a\times n_a}$ are diagonal positive definite matrices, representing the proportion and derivative gains, respectively.
$e(t)$ and $\dot{e}(t)$ are the tracking error on actuated joints at time $t$ defined as
\begin{align}
    e_a(t) &= \dq - \q \\
    \dot{e}_a(t) &= \dqd - \qd.
\end{align}
We incorporate the identified parameters into an inverse dynamics controller to show that these parameters can enhance tracking performance in controller design.
The desired trajectories for all actuated joints were chosen to be sine waves with a maximum velocity of $1.8$, $3$, and $5$ rad/s.
These were distinct trajectories from those used for identification and cross-validation.

\begin{figure*}[t]
    \centering
    \includegraphics[width=1.9\columnwidth]{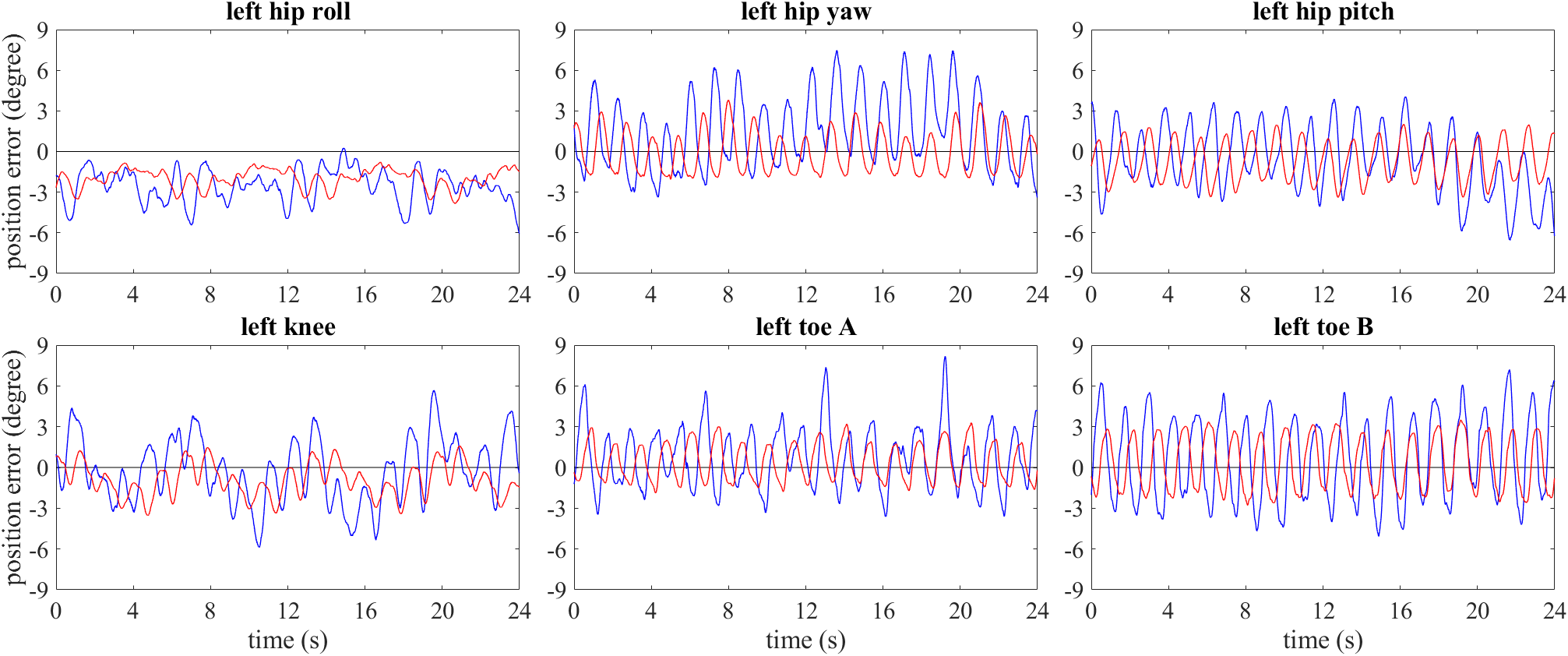}
    \caption{The plots show the tracking error of applying an inverse dynamics controller to track a desired trajectory of a sine wave with a maximum velocity of $5$ rad/s for all actuated joints on the left leg.
    We evaluate how the controller performs when integrated with identified parameters versus the manufacturer provided parameters.
    The \textcolor{red}{\textbf{tracking error corresponding to identified parameters}} is plotted in \textcolor{red}{\textbf{red}}.
    The \textcolor{blue}{\textbf{tracking error corresponding to manufacturer provided parameters}} is plotted in \textcolor{blue}{\textbf{blue}}.
    The desired trajectories shown are distinct from that used in identification.
    }
    \label{fig:tracking_error_results}
\end{figure*}

The tracking error for all actuated joints for the desired trajectories with a maximum velocity of $5$ rad/s can be found in Table \ref{tab:inverse_dynamics_controller_error}.
The tracking error plots are included in Figure \ref{fig:tracking_error_results}.
The tracking error using the parameters identified with our method is always maintained within 4 degrees for all actuated joints on legs and 1 degree for all actuated joints on arms.
We also show the tracking error corresponding to the manufacturer provided parameters in Table \ref{tab:inverse_dynamics_controller_error_comparison}, which is significantly higher than the tracking error corresponding to the identified parameters.
The tracking error results of the desired trajectories of 1.8 and 3 rad/s can be found in our online supplementary material \cite{supp}.

\begin{table}[t]
    \centering
    \begin{tabular}{c|c|c}
    \hline
    Actuated Joints & Left & Right \\
    \hline
    hip-roll  & (1.985, 3.848) & (0.917, 3.126) \\
    hip-yaw   & (1.398, 3.399) & (1.323, 3.807) \\
    hip-pitch & (1.343, 3.810) & (1.005, 2.045) \\
    knee      & (1.283, 3.546) & (1.135, 3.042) \\
    toe-A     & (1.217, 3.297) & (0.419, 2.869) \\
    toe-B     & (1.781, 3.492) & (0.679, 2.519) \\
    \hline
    shoulder-roll  & (0.172, 0.755) & (0.099, 0.481) \\
    shoulder-pitch & (0.317, 0.808) & (0.131, 0.612) \\
    shoulder-yaw   & (0.146, 0.547) & (0.106, 0.909) \\
    elbow          & (0.131, 0.746) & (0.144, 0.473) \\
    \hline
    \end{tabular}
    \caption{Tracking error of the inverse dynamics controller incorporated with \textbf{identified parameters} over all actuated joints for desired trajectories with a maximum speed of $5$ rad/s.
    We report the (mean, maximum) values of the tracking error in \textbf{degrees}.}
    \label{tab:inverse_dynamics_controller_error}
\end{table}

\begin{table}[t]
    \centering
    \begin{tabular}{c|c}
    \hline
    Actuated Joints & Left \\
    \hline
    hip-roll  & (2.805, 7.198) \\
    hip-yaw   & (2.537, 6.070) \\
    hip-pitch & (2.413, 7.471) \\
    knee      & (2.088, 6.559) \\
    toe-A     & (1.976, 5.887) \\
    toe-B     & (1.897, 8.181) \\
    \hline
    \end{tabular}
    \caption{Tracking error of the inverse dynamics controller incorporated with \textbf{manufacturer provided parameters} over all actuated joints for desired trajectories with a maximum speed of $5$ rad/s.
    We report the (mean, maximum) values of the tracking error in \textbf{degrees}.
    Note that the tracking error for all joints is significantly larger than the results in the "Left Leg" section in Table \ref{tab:inverse_dynamics_controller_error}.}
    \label{tab:inverse_dynamics_controller_error_comparison}
\end{table}
\section{Conclusion}
\label{sec:conclusion}

This paper proposes a novel approach for system identification for constrained systems.
The proposed method is able to identify physically realizable parameters using an optimization-based approach.
The algorithm is validated through extensive real-world experiments on Digit. 
Future work will consider how to perform system identification in the presence of ground contact.

\bibliographystyle{IEEEtran}
\bibliography{RAL2024SysID}

\begin{thebibliography}{10}
\providecommand{\url}[1]{#1}
\csname url@samestyle\endcsname
\providecommand{\newblock}{\relax}
\providecommand{\bibinfo}[2]{#2}
\providecommand{\BIBentrySTDinterwordspacing}{\spaceskip=0pt\relax}
\providecommand{\BIBentryALTinterwordstretchfactor}{4}
\providecommand{\BIBentryALTinterwordspacing}{\spaceskip=\fontdimen2\font plus
\BIBentryALTinterwordstretchfactor\fontdimen3\font minus \fontdimen4\font\relax}
\providecommand{\BIBforeignlanguage}[2]{{%
\expandafter\ifx\csname l@#1\endcsname\relax
\typeout{** WARNING: IEEEtran.bst: No hyphenation pattern has been}%
\typeout{** loaded for the language `#1'. Using the pattern for}%
\typeout{** the default language instead.}%
\else
\language=\csname l@#1\endcsname
\fi
#2}}
\providecommand{\BIBdecl}{\relax}
\BIBdecl

\bibitem{Gautier2011}
M.~Gautier, P.-O. Vandanjon, and A.~Janot, ``Dynamic identification of a 6 dof robot without joint position data,'' in \emph{2011 IEEE International Conference on Robotics and Automation}.\hskip 1em plus 0.5em minus 0.4em\relax IEEE, May 2011.

\bibitem{Han2020}
Y.~Han, J.~Wu, C.~Liu, and Z.~Xiong, ``An iterative approach for accurate dynamic model identification of industrial robots,'' \emph{IEEE Transactions on Robotics}, vol.~36, no.~5, pp. 1577--1594, Oct. 2020.

\bibitem{Janot2014AGI}
\BIBentryALTinterwordspacing
A.~Janot, P.-O. Vandanjon, and M.~Gautier, ``A generic instrumental variable approach for industrial robot identification,'' \emph{IEEE Transactions on Control Systems Technology}, vol.~22, pp. 132--145, 2014. [Online]. Available: \url{https://api.semanticscholar.org/CorpusID:23721529}
\BIBentrySTDinterwordspacing

\bibitem{paper-agility}
\BIBentryALTinterwordspacing
J.~Hurst, ``Building robots that can go where we go,'' in \emph{IEEE Spectrum: Technology, Engineering, and Science News}, 2019. [Online]. Available: \url{https://spectrum.ieee.org/robotics/humanoids/building-robots-that-can-go-where-we-go}
\BIBentrySTDinterwordspacing

\bibitem{paper-atlas}
\BIBentryALTinterwordspacing
B.~Dynamics. (2024) Atlas and beyond: the world's most dynamic robots. [Online]. Available: \url{https://bostondynamics.com/atlas/}
\BIBentrySTDinterwordspacing

\bibitem{paper-teslabot}
\BIBentryALTinterwordspacing
T.~A.~. Robotics. (2024) Tesla bot. [Online]. Available: \url{https://www.tesla.com/AI}
\BIBentrySTDinterwordspacing

\bibitem{An1985}
C.~An, C.~Atkeson, and J.~Hollerbach, ``Estimation of inertial parameters of rigid body links of manipulators,'' in \emph{1985 24th IEEE Conference on Decision and Control}.\hskip 1em plus 0.5em minus 0.4em\relax IEEE, Dec. 1985.

\bibitem{Gautiera}
M.~Gautier, ``Numerical calculation of the base inertial parameters of robots,'' in \emph{Proceedings., IEEE International Conference on Robotics and Automation}, 1990, pp. 1020--1025 vol.2.

\bibitem{Gautier1990}
M.~Gautier and W.~Khalil, ``Direct calculation of minimum set of inertial parameters of serial robots,'' \emph{IEEE Transactions on Robotics and Automation}, vol.~6, no.~3, pp. 368--373, Jun. 1990.

\bibitem{Khalil1994}
W.~Khalil and F.~Bennis, ``Comments on “direct calculation of minimum set of inertial parameters of serial robots”,'' \emph{IEEE Transactions on Robotics and Automation}, vol.~10, no.~1, pp. 78--79, Feb. 1994.

\bibitem{Venture2009}
G.~Venture, K.~Ayusawa, and Y.~Nakamura, ``A numerical method for choosing motions with optimal excitation properties for identification of biped dynamics - an application to human,'' in \emph{2009 IEEE International Conference on Robotics and Automation}.\hskip 1em plus 0.5em minus 0.4em\relax IEEE, May 2009.

\bibitem{Sousa2014}
C.~D. Sousa and R.~Cortesão, ``Physical feasibility of robot base inertial parameter identification: A linear matrix inequality approach,'' \emph{The International Journal of Robotics Research}, vol.~33, no.~6, pp. 931--944, Feb. 2014.

\bibitem{Sousa2019}
C.~D. Sousa and R.~Cortesao, ``Inertia tensor properties in robot dynamics identification: A linear matrix inequality approach,'' \emph{IEEE/ASME Transactions on Mechatronics}, vol.~24, no.~1, pp. 406--411, Feb. 2019.

\bibitem{Wensing2018}
P.~M. Wensing, S.~Kim, and J.-J.~E. Slotine, ``Linear matrix inequalities for physically consistent inertial parameter identification: A statistical perspective on the mass distribution,'' \emph{IEEE Robotics and Automation Letters}, vol.~3, no.~1, pp. 60--67, Jan. 2018.

\bibitem{paper-tree-structure}
W.~Khalil, F.~Bennis, and M.~Gautier, ``Calculation of the minimum inertial parameters of tree structure robots,'' in \emph{Advanced Robotics: 1989: Proceedings of the 4th International Conference on Advanced Robotics Columbus, Ohio, June 13--15, 1989}.\hskip 1em plus 0.5em minus 0.4em\relax Springer, 1989, pp. 189--201.

\bibitem{Bennis1992CalculationOT}
\BIBentryALTinterwordspacing
F.~Bennis, W.~Khalil, and M.~Gautier, ``Calculation of the base inertial parameters of closed-loops robots,'' \emph{Proceedings 1992 IEEE International Conference on Robotics and Automation}, pp. 370--375 vol.1, 1992. [Online]. Available: \url{https://api.semanticscholar.org/CorpusID:32387963}
\BIBentrySTDinterwordspacing

\bibitem{Tan2022}
C.~Tan, H.~Zhao, and H.~Ding, ``Identification of dynamic parameters of closed-chain industrial robots considering motor couplings,'' \emph{Computers and Electrical Engineering}, vol.~99, p. 107740, Apr. 2022.

\bibitem{Shome_1998}
S.~S. Shome, D.~G. Beale, and D.~Wang, ``A general method for estimating dynamic parameters of spatial mechanisms,'' \emph{Nonlinear Dynamics}, vol.~16, no.~4, pp. 349--368, 1998.

\bibitem{featherstone2014rigid}
R.~Featherstone, \emph{Rigid body dynamics algorithms}.\hskip 1em plus 0.5em minus 0.4em\relax Springer, 2014.

\bibitem{paper-regressor}
H.~Mayeda, K.~Yoshida, and K.~Osuka, ``Base parameters of manipulator dynamic models,'' in \emph{Proceedings. 1988 IEEE International Conference on Robotics and Automation}.\hskip 1em plus 0.5em minus 0.4em\relax IEEE, 1988, pp. 1367--1372.

\bibitem{JerryShurman}
J.~Shurman, \emph{Multivariable Calculus}.\hskip 1em plus 0.5em minus 0.4em\relax Reed College, 2014.

\bibitem{supp}
``Online supplementary information for system identification for constrained robots,'' \url{https://github.com/roahmlab/ConstrainedSysID/blob/main/Supplementary_Materials/Contrained_SysID_Appendix.pdf}, accessed: 2024-08-16.

\bibitem{paper-unitree}
\BIBentryALTinterwordspacing
U.~Robotics. (2024) Unitree's first universal humanoid robot. [Online]. Available: \url{https://www.unitree.com/cn/h1/}
\BIBentrySTDinterwordspacing

\bibitem{paper-fourier-intelligence}
\BIBentryALTinterwordspacing
F.~Intelligence. (2024) Fourier gr1. [Online]. Available: \url{https://fourierintelligence.com/gr1/}
\BIBentrySTDinterwordspacing

\bibitem{paper-agility-v4}
\BIBentryALTinterwordspacing
A.~Robotics, ``The next generation of digit - enabling humans to be more human,'' in \emph{Youtube}, 2023. [Online]. Available: \url{https://www.youtube.com/watch?v=rnFZAB9ogEE}
\BIBentrySTDinterwordspacing

\bibitem{paper-ode15s}
L.~F. Shampine and M.~W. Reichelt, ``The matlab ode suite,'' \emph{SIAM Journal on Scientific Computing}, vol.~18, pp. 1--22, 1997.

\end{thebibliography}

\appendices
\section{Derivation  of Theorem \ref{thm-fully-actuated}} 
\label{app-thm-fully-actuated}

The following proof is modified from \cite[Section 8.8, 8.11, 8.12]{featherstone2014rigid}, with a more specific formulation of $G(\q)$.
Readers are recommended to refer to \cite[Section 8]{featherstone2014rigid} for a more comprehensive introduction to systems with closed-loop kinematic constraints.

For the sake of simplicity, we remove the time notation in this proof.
We abuse the notation $q$ and define it as a point in the robot configuration space $\Q$.
The inertia matrix, the Coriolis matrix, and the gravity vector are then written as
$H(q)$, $C(q, \dot{q})$, and $g(q)$, respectively.
Hence, the dynamics equations are written as
\begin{equation} \label{eq-unconstrained-dynamics}
    Y(q, \dot{q}, \ddot{q})\pi = Bu + J^T(q)\lambda.
\end{equation}

Given Assumption \ref{assum-uniqueIK}, according to the inverse function theorem,
for $\forall q_u\in\Q_u$ and $q_a\in\Q_a$, s.t. $c(q) = 0$, $J_u(q)\in\R^{n_u\times n_u}$ is invertible.

\subsection{Derivation of \eqref{eq-fill-full-joints1} and \eqref{eq-fill-full-joints2}}
The constraints are satisfied all the time, which means that the time derivative and the second-order time derivative are also zero.
We first decompose \eqref{eq:cons-vel} into unactuated part and actuated part:
\begin{align}
    \begin{split}
        \dot{c}(q) &= J(q)\dot{q} \\
                   &= J_u(q)\dot{q}_u + J_a(q)\dot{q}_a = 0.
    \end{split} 
\end{align}
Solving for $\dot{q}_u$, we get
\begin{equation}
    \dot{q}_u = -J_u^{-1}(q)J_a(q)\dot{q}_a.
\end{equation}

The transform matrix $G(q)\in\R^{n\times n_a}$ is defined as
\begin{equation}
    G_u(q) = -J_u^{-1}(q) J_a(q),\ G_a(q) = \mathbb{1}_{n_a \times n_a},
\end{equation}
where $G_u(q)$ and $G_a(q)$ are the unactuated columns and the actuated columns of $G(q)$, respectively.
Hence,
\begin{equation}
    \dot{q}_u = G_u(q)\dot{q_a},\ \dot{q}_a = G_a(q)\dot{q_a}.
\end{equation}
As a result, we can get \eqref{eq-fill-full-joints1} by concatenating two components above:
\begin{equation} \label{eq:projection-vel}
    \dot{q} = G(q)\dot{q}_a. 
\end{equation}
By differentiate \eqref{eq:projection-vel} with time, we can get \eqref{eq-fill-full-joints2}.

\subsection{Derivation of \eqref{eq-constrainedID-torque}}

We first denote the uncontrained dynamics vector $Y(q, \dot{q}, \ddot{q})\pi$ using $\Tilde{u}$.
\eqref{eq-unconstrained-dynamics} then can be simplified to
\begin{equation}
    \Tilde{u} = Bu + J^T(q)\lambda
\end{equation}


To get \eqref{eq-constrainedID-torque}, first multiply $G^T(q)$ to both sides of \eqref{eq-unconstrained-dynamics}:
\begin{equation} \label{eq:projection-dynamics}
    G^T(q)\Tilde{u} = G^T(q)Bu + G^T(q)J^T(q)\lambda.
\end{equation}
For the right side of \eqref{eq:projection-dynamics},
since
\begin{align*}
    B_u &= \mathbb{0}_{n_u\times n_a} \\
    B_a &= \mathbb{1}_{n_a\times n_a},
\end{align*}
$G^T(q)Bu$ can be simplified:
\begin{subequations}
\begin{align}
    G^T(q)Bu &= (G_a^T(q)B_a + G_u^T(q)B_u)u \\
    &=G_a^T(q)B_au\\
    &= \mathbb{1}_{n_a \times n_a}u \\
    &= u.
\end{align}
\end{subequations}
$G^T(q)J^T(q)\lambda$ can be simplified:
\begin{subequations}
\begin{align}
    G^T(q)J^T(q)\lambda &= (J(q)G(q))^T\lambda \\
                 &= (J_a(q)G_a(q) + J_u(q)G_u(q))^T\lambda \\
                 &= (J_a(q) - J_u(q)J_u^{-1}(q)J_a(q))^T\lambda \\
                 &= (J_a(q) - J_a(q))^T\lambda \\
                 &= \mathbb{0}_{n_a\times n_u}\lambda \\
                 &= \mathbb{0}_{n_a}.
\end{align}
\end{subequations}
Hence \eqref{eq:projection-dynamics} is finally simplified to
\begin{align} \label{eq-actuated-space-dynamics-simplified}
    u &= G^T(q)\Tilde{u} \\
    &= G^T(q)Y(q, \dot{q}, \ddot{q})\pi,
\end{align}
which gives us the same equation in \eqref{eq-constrainedID-torque}.
\qed

\end{document}